\documentclass[preprint,12pt]{elsarticle}

\usepackage{amsmath,amssymb,amsfonts}
\usepackage{graphicx}
\usepackage{subcaption} 
\usepackage{tikz}
\usetikzlibrary{
	positioning,
	arrows.meta,
	shapes.geometric,
	fit,
	shadows.blur,
	calc
}
\usepackage{booktabs} 
\usepackage{siunitx} 
\usepackage{hyperref} 
\usepackage{xcolor} 
\usepackage{multirow} 
\usepackage{float} 
\usepackage{enumitem} 
\usepackage{csvsimple} 
\usepackage{pgfplotstable} 
\usepackage{xstring}

\definecolor{encoderblock}{HTML}{E0F2F7}
\definecolor{decoderblock}{HTML}{FFF3E0}
\definecolor{skipconnection}{HTML}{A7D9B5}
\definecolor{arrowcolor}{HTML}{424242}
\definecolor{mathcolor}{HTML}{0056B3}

\journal{Biomedical Signal Processing and Control}

\begin{document}
	
	\begin{frontmatter}
		
		\title{ConvNeXt-FD: A Fractal-Based Deep Model for Robust Biomedical Image Segmentation}
		
		\author[inst1]{Joao Batista Florindo}
		\author[inst1]{Amanda Pontes de Oliveira Ornelas}
		\affiliation[inst1]{organization={Institute of Mathematics, Statistics and Scientific Computing, Department of Applied Mathematics, University of Campinas},
			addressline={Rua Sergio Buarque de Holanda, 651},
			city={Campinas},
			postcode={13083-859},
			state={Sao Paulo},
			country={Brazil}}
		
		\begin{abstract}
			Biomedical image segmentation is a critical task in medical diagnosis and treatment planning, enabling precise delineation of anatomical structures and pathological regions. Despite significant advancements, challenges persist due to the inherent variability, noise, and complex morphology present in diverse medical imaging modalities. This paper introduces ConvNeXt-FD, a novel deep learning architecture for robust biomedical image segmentation, built upon a U-Net-like encoder-decoder framework leveraging the powerful ConvNeXt backbone. Our approach integrates a hybrid loss function combining the Dice coefficient with a boundary-aware regularization term inspired by a differentiable formulation of Fractal Dimension, designed to enhance the model's sensitivity to object boundaries and shape fidelity. We rigorously evaluate ConvNeXt-FD across six distinct biomedical datasets: BUSI (Breast Ultrasound Images), DDTI (Thyroid Ultrasound Images), FluoCells (Fluorescent Cell Images), IDRiD (Diabetic Retinopathy Images for Optic Disc Segmentation), ISIC2018 (Skin Lesion Images), and MoNuSeg (Nuclei Segmentation). Experimental results demonstrate that ConvNeXt-FD, particularly when initialized with ImageNet pre-trained weights, achieves competitive and often superior performance compared to existing state-of-the-art methods across various metrics, including Dice, Jaccard, Accuracy, Sensitivity, Specificity, and False Positive Rate. The integration of ConvNeXt as a strong encoder, coupled with the boundary-aware regularization, proves effective in capturing both high-level semantic features and fine-grained boundary details, leading to more accurate and reliable segmentations in challenging biomedical contexts.
		\end{abstract}
		
		
	\end{frontmatter}
	
	\section{Introduction}
	Biomedical image segmentation plays a pivotal role in numerous clinical applications, ranging from tumor detection and organ delineation to cell analysis and disease progression monitoring \cite{gao2025c}. Accurate segmentation provides quantitative insights crucial for diagnosis, treatment planning, and surgical guidance. However, the inherent complexities of medical images, such as low contrast, intensity inhomogeneities, artifacts, and significant inter-patient variability, pose substantial challenges for automated segmentation systems.
	
	Deep learning, particularly Convolutional Neural Networks (CNNs), has revolutionized medical image analysis. The U-Net architecture \cite{ronneberger2015u}, with its symmetric encoder-decoder structure and skip connections, has become a \textit{de facto} standard for biomedical image segmentation due to its ability to capture both contextual information and fine-grained details. Numerous variants and extensions of U-Net have been proposed, incorporating attention mechanisms, residual connections, and multi-scale feature fusion to further enhance performance.
	
	More recently, Vision Transformers (ViTs), State-Space Models (Mamba), and their hybrid counterparts have emerged, demonstrating impressive capabilities in capturing long-range dependencies, which are often crucial in complex medical images \cite{chen2024a, ruan2024b, xing2401}. Concurrently, modern CNN architectures like ConvNeXt \cite{liu2022convnet} have re-evaluated traditional CNN design principles, demonstrating that carefully modernized ConvNets can achieve performance comparable to or even surpassing Transformers, while retaining the inductive biases beneficial for image processing.
	
	Despite these advancements, achieving robust and precise segmentation, especially at object boundaries, remains an active area of research. Traditional loss functions, such as binary cross-entropy (BCE) and Dice loss, often struggle with small objects, imbalanced classes, and fuzzy boundaries. To address these limitations, researchers have explored various boundary-aware loss functions and regularization techniques.
	
	In this paper, we propose ConvNeXt-FD, a novel deep learning framework for biomedical image segmentation. Our architecture is based on a U-Net-like design, employing a ConvNeXt model as its powerful encoder. To further improve segmentation quality, particularly at object boundaries, we introduce a hybrid loss function that combines the widely used Dice loss with a boundary-aware regularization term. This regularization term is inspired by the principles of Fractal Dimension \cite{cannon1984fractal,jabdaragh2023mtfd}, aiming to minimize the discrepancy between the predicted and ground truth object shapes. We hypothesize that by explicitly encouraging shape fidelity through this boundary-aware regularization, ConvNeXt-FD can achieve more accurate and clinically relevant segmentations across a diverse range of biomedical imaging tasks.
	
	We evaluate ConvNeXt-FD on six challenging and diverse public biomedical datasets: Breast Ultrasound Images (BUSI), Diabetic Retinopathy Images (IDRiD) for optic disc segmentation, Thyroid Ultrasound Images (DDTI), Fluorescent Cell Images (FluoCells), Skin Lesion Images (ISIC2018), and Nuclei Segmentation (MoNuSeg). Our comprehensive experimental analysis demonstrates the effectiveness of the proposed architecture and loss function, showcasing its ability to achieve state-of-the-art performance across these varied tasks.
	
	The remainder of this paper is organized as follows: Section \ref{sec:related_works} reviews recent advancements in medical image segmentation. Section \ref{sec:proposed_method} details the ConvNeXt-FD architecture and its mathematical formulation. Section \ref{sec:experimental_setup} describes the experimental protocols, datasets, and evaluation metrics. Section \ref{sec:results} presents and analyzes our empirical results. Section \ref{sec:sota_comparison} compares our findings with the current state-of-the-art. Finally, Section \ref{sec:conclusion} concludes the paper and outlines future research directions.
	
	\section{Related Works}
	\label{sec:related_works}
	Deep learning has profoundly impacted biomedical image segmentation, with the U-Net architecture \cite{ronneberger2015u} serving as a foundational model. Its success stems from its ability to integrate multi-scale features through skip connections, effectively capturing both high-level semantic context and low-level spatial details. Numerous variants have since been developed to address specific challenges in medical imaging.
	
	Early advancements focused on enhancing U-Net's core components. U-Net++ \cite{zhou2020unet} introduced nested and dense skip connections to improve feature aggregation at different scales. Attention U-Net \cite{oktay2018attention} incorporated attention gates to highlight salient features and suppress irrelevant background noise, leading to more robust segmentations. Residual connections, as seen in ResUNet \cite{zhang2018road}, were adopted to facilitate deeper network training and mitigate vanishing gradients.
	
	The advent of Transformer architectures in computer vision has opened new avenues for medical image segmentation. Models like TransUNet \cite{chen2021transunet} and Swin-UNet \cite{cao2021swin} combine the strengths of CNNs for local feature extraction with Transformers' ability to model long-range dependencies. Recent works continue to explore the integration of Transformers, with architectures like VM-UNet \cite{ruan2024b} and SegMamba \cite{xing2401} leveraging state space models (Mamba) for efficient sequential modeling in medical image segmentation. Other approaches, such as Zig-RiR \cite{chen2025h}, explore novel recurrent neural network architectures for efficiency.
	
	Parallel to Transformer advancements, modern CNNs have undergone a resurgence. Architectures like ConvNeXt \cite{liu2022convnet} demonstrate that by re-evaluating design choices from the Transformer literature (e.g., larger kernel sizes, inverted bottleneck blocks, layer normalization), pure ConvNets can achieve competitive performance. This has led to a renewed interest in optimizing CNNs for medical tasks, as they often offer better computational efficiency and retain strong inductive biases beneficial for image data.
	
	Loss functions also play a critical role in segmentation performance. Beyond standard Binary Cross-Entropy (BCE) and Dice Loss \cite{milletari2016v}, specialized losses have been developed to handle class imbalance (e.g., Focal Loss \cite{lin2017focal}), boundary delineation (e.g., Hausdorff Distance Loss \cite{karimi2019image}), and shape fidelity. For instance, some methods focus on learning from AI-generated annotations \cite{song2025e} or adapting vision foundation models \cite{li2025i}. The concept of Fractal Dimension, traditionally used to measure the irregularity/complexity of curves or shapes, inspires our boundary-aware regularization, aiming to improve the geometric accuracy of segmentations.
	
	Recent literature also highlights the importance of semi-supervised learning \cite{zeng2025d} and enforcing attention usage \cite{wald2025g} to improve model generalization and performance with limited labeled data. Comprehensive reviews \cite{gao2025c, wang2025j} provide valuable insights into the landscape of deep learning methods for medical image segmentation, emphasizing the continuous evolution of architectures and techniques. Our work builds upon these foundations by integrating a robust ConvNeXt encoder with a novel Fractal-inspired boundary regularization, aiming for superior performance across diverse biomedical segmentation tasks.
	
	\section{Proposed Method: ConvNeXt-FD}
	\label{sec:proposed_method}
	Our proposed architecture, ConvNeXt-FD, is designed for robust biomedical image segmentation by combining the strengths of a modern Convolutional Neural Network (CNN) backbone with a specialized hybrid loss function. The architecture follows a U-Net-like encoder-decoder paradigm, where the encoder is built using ConvNeXt blocks, and the decoder reconstructs the segmentation mask by progressively upsampling features and integrating information from the encoder via skip connections. A key innovation lies in our hybrid loss function, which incorporates a boundary-aware regularization term inspired by the Fractal Dimension to enhance shape fidelity.
	
	\subsection{Architectural Design}
\begin{figure}[!htpb]
		\centering

			\begin{tikzpicture}[
				font=\small,
				>=Latex,
				node distance=1.4cm and 1.6cm,				
				block/.style={
					rectangle,
					rounded corners=4pt,
					draw=black,
					thick,
					align=center,
					minimum width=3.4cm,
					minimum height=1.1cm,
					fill=blue!10,
					blur shadow
				},				
				process/.style={
					rectangle,
					rounded corners=4pt,
					draw=black,
					thick,
					align=center,
					minimum width=3.8cm,
					minimum height=1.1cm,
					fill=green!12,
					blur shadow
				},		
				loss/.style={
					rectangle,
					rounded corners=4pt,
					draw=black,
					thick,
					align=center,
					minimum width=3.8cm,
					minimum height=1.1cm,
					fill=red!10,
					blur shadow
				},			
				output/.style={
					rectangle,
					rounded corners=4pt,
					draw=black,
					thick,
					align=center,
					minimum width=3.2cm,
					minimum height=1.1cm,
					fill=orange!15,
					blur shadow
				},				
				arrow/.style={
					->,
					thick
				}
				]
				
%
%
%
				
				
				\node[process] (encoder) {
					ConvNeXt-Large Encoder\\
					(ImageNet Pretrained)
				};
				
				\node[process, below=of encoder] (decoder) {
					U-Net Decoder
				};
				
				\node[output, below left=1.8cm and 1.8cm of decoder] (seghead) {
					Segmentation Head\\
					Binary Mask
				};
				
				\node[output, below right=1.8cm and 1.8cm of decoder] (fdhead) {
					Fractal Dimension Head\\
					FD Map
				};
				
				
				\node[process, right=2cm of seghead] (fdtarget) {
					Ground Truth FD Map\\
					Edge Extraction\\
					Laplacian Kernel
				};
				
				
				\node[loss, below=2cm of seghead] (segloss) {
					Segmentation Loss\\
					BCE + Dice
				};
				
				\node[loss, below=2cm of fdhead] (fdloss) {
					Fractal Loss\\
					MSE
				};
				
				\node[loss, below=2.2cm of $(segloss)!0.5!(fdloss)$] (total) {
					Total ConvNeXt-FD Loss\\
					$\mathcal{L} =
					\mathcal{L}_{seg}
					+ \lambda_{fd}\mathcal{L}_{fd}$
				};
				
				
%
				
				
%
				\draw[arrow] (encoder) -- (decoder);
				
				\draw[arrow] (decoder) -- (seghead);
				\draw[arrow] (decoder) -- (fdhead);
				
				\draw[arrow] (seghead) -- (segloss);
				\draw[arrow] (fdhead) -- (fdloss);
				
				\draw[arrow] (fdtarget) -- (fdloss);
				
				\draw[arrow] (segloss) -- (total);
				\draw[arrow] (fdloss) -- (total);
				
				
				
				\node[
				draw=black!60,
				dashed,
				rounded corners=8pt,
				inner sep=10pt,
				fit=(encoder)(decoder)(seghead)(fdhead),
				label={[yshift=0.4cm]north:\bfseries ConvNeXt-FD Architecture}
				] {};
				
			\end{tikzpicture}
		\caption{Architectural overview of ConvNeXt-FD. The encoder path utilizes ConvNeXt blocks for hierarchical feature extraction, followed by a bottleneck. The decoder path reconstructs the segmentation mask through upsampling and concatenation with skip connections from the encoder. The final output is optimized using a hybrid loss function incorporating a boundary-aware term.}
		\label{fig:architecture}
	\end{figure}
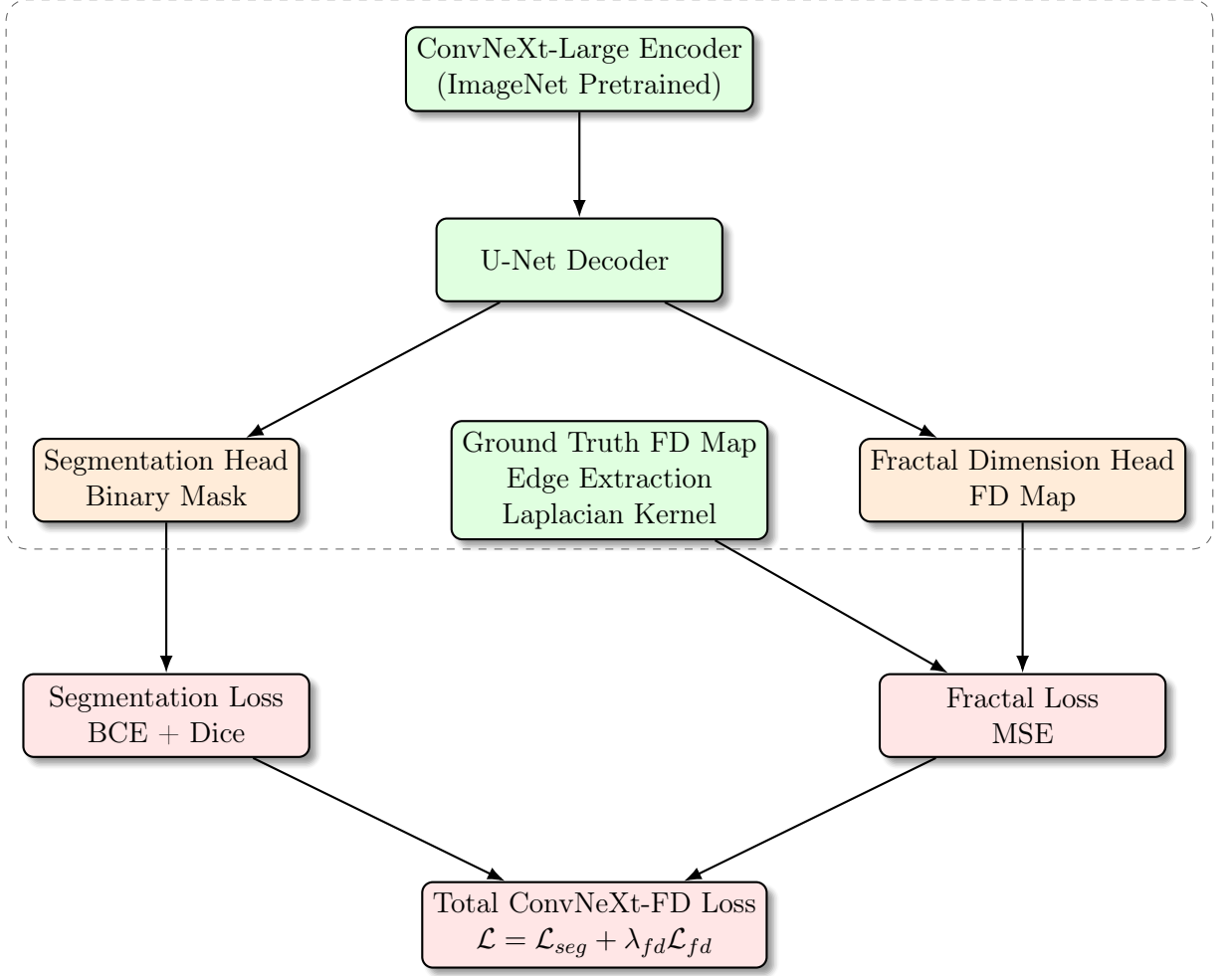

	The ConvNeXt-FD architecture, depicted in Figure \ref{fig:architecture}, comprises an encoder, a bottleneck, and a decoder.
	
	\subsubsection{ConvNeXt Encoder}
	The encoder is responsible for extracting hierarchical features from the input image $X \in \mathbb{R}^{H \times W \times C}$, where $H, W$ are height and width, and $C$ is the number of channels (e.g., $C=3$ for RGB images). We adopt the ConvNeXt architecture \cite{liu2022convnet} as our encoder due to its strong performance and efficient design, which incorporates modern CNN principles.
	
	A ConvNeXt encoder typically consists of a stem layer followed by several stages. Each stage $s \in \{0, \dots, S-1\}$ applies a series of ConvNeXt blocks, which are characterized by:
	\begin{itemize}
		\item \textbf{Large Kernel Depthwise Convolution:} A $7 \times 7$ depthwise convolution, $D_k$, is applied, followed by Layer Normalization (LN) and a $1 \times 1$ convolution. This is a departure from traditional inverted bottlenecks.
		\item \textbf{Inverted Bottleneck Structure:} Similar to Transformers' MLP blocks, a $1 \times 1$ convolution expands the channel dimension, followed by a GELU activation, another $1 \times 1$ convolution to project back to the original dimension, and a residual connection.
		\item \textbf{Layer Normalization:} Applied before convolution layers, enhancing stability.
	\end{itemize}
	Let $F_s(X)$ denote the feature map at stage $s$. The operation of a single ConvNeXt block $B_s$ at stage $s$ can be formally described as:
	\begin{equation}
		Z_s = D_k(F_s)
	\end{equation}
	\begin{equation}
		\tilde{Z}_s = \text{LN}(Z_s)
	\end{equation}
	\begin{equation}
		\hat{Z}_s = \text{Conv}_{1 \times 1}(\tilde{Z}_s)
	\end{equation}
	\begin{equation}
		\bar{Z}_s = \text{GELU}(\hat{Z}_s)
	\end{equation}
	\begin{equation}
		F_{s+1} = \text{Conv}_{1 \times 1}(\bar{Z}_s) + F_s
	\end{equation}
	where $D_k$ is the depthwise convolution with kernel size $k \times k$ (typically $k=7$), $\text{LN}$ is Layer Normalization, $\text{Conv}_{1 \times 1}$ denotes a $1 \times 1$ convolution, and $\text{GELU}$ is the Gaussian Error Linear Unit activation function. Downsampling between stages is achieved using $2 \times 2$ convolutions with stride 2, reducing spatial dimensions while increasing channel depth. The encoder produces a set of feature maps $\{E_1, E_2, E_3, E_4, E_5\}$ at progressively smaller spatial resolutions and higher channel depths.
	
	\subsubsection{Decoder and Skip Connections}
	The decoder path takes the deepest feature map from the encoder (the bottleneck output) and progressively upsamples it to reconstruct the segmentation mask. At each decoding stage $d \in \{1, \dots, 5\}$, the upsampled feature map $U_d$ is concatenated with the corresponding skip connection $E_{5-d+1}$ from the encoder path. This concatenation allows the decoder to leverage both high-level semantic information from deeper layers and fine-grained spatial details from shallower layers.
	\begin{equation}
		D_d = \text{Concat}(\text{Upsample}(D_{d-1}), E_{5-d+1})
	\end{equation}
	where $D_0$ is the bottleneck output, and $\text{Upsample}$ typically involves a transposed convolution or bilinear interpolation followed by a convolution. The concatenated features are then processed by a series of convolutional layers to refine the segmentation. The final layer of the decoder applies a $1 \times 1$ convolution followed by a sigmoid activation function to produce the predicted segmentation mask $\hat{Y} \in [0, 1]^{H \times W \times 1}$.
	
	\subsection{Hybrid Loss Function: Fractal Dimension-Regularized Loss}
	\label{sec:loss_function}
	Accurate segmentation of biomedical images often requires not only high pixel-wise accuracy but also precise delineation of object boundaries. To achieve this, we propose a hybrid loss function $\mathcal{L}$ that combines a standard segmentation loss with a boundary-aware regularization term.
	\begin{equation}
		\mathcal{L}(Y, \hat{Y}) = \mathcal{L}_{\text{Dice}}(Y, \hat{Y}) + \lambda_{\text{FD}} \mathcal{L}_{\text{Boundary}}(Y, \hat{Y})
		\label{eq:hybrid_loss}
	\end{equation}
	where $Y$ is the ground truth mask, $\hat{Y}$ is the predicted mask, and $\lambda_{\text{FD}}$ is a weighting hyperparameter controlling the influence of the boundary-aware term.
	
	\subsubsection{Dice Loss}
	The Dice Loss ($\mathcal{L}_{\text{Dice}}$) is a widely used metric for medical image segmentation, particularly effective for imbalanced datasets. It is derived from the Dice coefficient, which measures the overlap between two segmentation masks. For binary segmentation, the Dice coefficient $D$ is defined as:
	\begin{equation}
		D(Y, \hat{Y}) = \frac{2 |Y \cap \hat{Y}|}{|Y| + |\hat{Y}|} = \frac{2 \sum_{i=1}^{N} y_i \hat{y}_i}{\sum_{i=1}^{N} y_i + \sum_{i=1}^{N} \hat{y}_i}
	\end{equation}
	where $y_i$ and $\hat{y}_i$ are the pixel values of the ground truth and predicted masks, respectively, and $N$ is the total number of pixels. The Dice Loss is then formulated as:
	\begin{equation}
		\mathcal{L}_{\text{Dice}}(Y, \hat{Y}) = 1 - D(Y, \hat{Y})
	\end{equation}
	To ensure differentiability, especially with sigmoid outputs, a small smoothing term $\epsilon$ is often added to the numerator and denominator.
	
	\subsubsection{Boundary-Aware Fractal-Inspired Loss ($\mathcal{L}_{\text{Boundary}}$)}
	
	Handling complex texture variations and precise boundary delineation are crucial in many biomedical applications. Inspired by fractal geometry, which effectively quantifies texture complexity through self-similar patterns, we introduce an auxiliary Fractal Dimension (FD) learning task. Unlike Euclidean metrics, fractal dimensions can effectively model the irregularity of textures and rugged surfaces where boundaries might otherwise be indiscernible.
	
	Following the differential box-counting method, we transform the original input image $I$ into a ground truth FD map. For a pixel with coordinates $(x,y)$ and a kernel $w_r(x,y)$ of size $r \times r$, the fractal dimension is estimated within an $R \times R$ neighborhood. First, for scaling factors $2 \le r \le R$, the number of boxes $N_r(x,y)$ is calculated based on the maximum and minimum pixel intensities within the kernel:
	\begin{equation}
		N_r(x,y) = \frac{R^2}{r^2} \left( \left[ \frac{M_r(x,y) - m_r(x,y)}{r} \right] + 1 \right)
	\end{equation}
	where the local extrema are defined as:
	\begin{align}
		M_r(x,y) &= \max_{(u,v) \in w_r(x,y)} I(u,v) \\
		m_r(x,y) &= \min_{(u,v) \in w_r(x,y)} I(u,v)
	\end{align}
	The fractal dimension $FD(x,y)$ for the pixel is then estimated as the slope of the least-squares regression line plotted on the points $(\log(1/r), \log(N_r(x,y)))$. 
	
	Because the fractal dimensions of boundaries between different types of regions exhibit high contrast differences, learning this FD map provides strong auxiliary supervision to the main segmentation task. We define the Fractal Dimension loss, $\mathcal{L}_{\text{Boundary}}$, as the Mean Squared Error (MSE) between the ground truth FD map computed from the image and the FD map predicted by the auxiliary decoder branch, $\widehat{FD}$:
	\begin{equation}
		\mathcal{L}_{\text{Boundary}} = \frac{1}{N} \sum_{i=1}^{N} \left( FD(p_i) - \widehat{FD}(p_i) \right)^2
	\end{equation}
	where $N$ is the total number of pixels. The hyperparameter $\lambda_{\text{FD}}$ in Equation \ref{eq:hybrid_loss} controls the contribution of this fractal dimension regularization, enforcing the network to effectively incorporate texture complexity information into the segmentation process.

	\section{Experimental Setup}
	\label{sec:experimental_setup}
	To thoroughly evaluate the performance of ConvNeXt-FD, we conducted experiments on six diverse biomedical image segmentation datasets.
	
	\subsection{Datasets}
	\begin{itemize}[leftmargin=*,labelsep=5mm]
		\item \textbf{BUSI (Breast Ultrasound Images)}: This dataset contains breast ultrasound images with corresponding ground truth masks for tumor segmentation. It is crucial for early detection and diagnosis of breast cancer. The dataset exhibits variability in image quality, lesion size, and shape.
		\item \textbf{DDTI (Thyroid Ultrasound Images)}: Focused on thyroid nodule segmentation from ultrasound images. Accurate delineation of thyroid nodules is vital for assessing malignancy risk. This dataset presents challenges related to subtle nodule boundaries and varying tissue textures.
		\item \textbf{FluoCells (Fluorescent Cell Images)}: Comprises fluorescent microscopy images of cells, with masks for individual cell segmentation. This is a fundamental task in biological research, often complicated by cell clustering and irregular shapes.
		\item \textbf{IDRiD (Indian Diabetic Retinopathy Image Dataset)}: Specifically, we used the optic disc segmentation task from IDRiD. The optic disc is a key anatomical landmark in retinal fundus images, and its precise segmentation is important for diagnosing and monitoring various ocular diseases, including glaucoma and diabetic retinopathy.
		\item \textbf{ISIC2018 (Skin Lesion Images)}: Part of the International Skin Imaging Collaboration (ISIC) challenge, this dataset provides dermoscopic images of skin lesions with corresponding segmentation masks. Accurate segmentation is a prerequisite for melanoma detection and classification.
		\item \textbf{MoNuSeg (Multi-Organ Nuclei Segmentation)}: It features highly varied tissue images across multiple organs, presenting significant morphological diversity. A primary challenge of this dataset is the need to accurately distinguish individual nuclei instances that are often tightly packed within dense clusters.
	\end{itemize}
	For each dataset, images and masks were preprocessed by resizing to a uniform dimension (e.g., $256 \times 256$ pixels) and normalizing pixel intensities to the range $[0, 1]$. Standard data augmentation techniques, including random rotations, horizontal and vertical flips, and brightness adjustments, were applied during training to enhance model generalization.
	
	\subsection{Evaluation Metrics}
	The performance of our segmentation model was quantitatively assessed using several widely accepted metrics:
	\begin{itemize}[leftmargin=*,labelsep=5mm]
		\item \textbf{Dice Coefficient (DSC)}: Measures the overlap between the predicted segmentation mask $\hat{Y}$ and the ground truth mask $Y$.
		\begin{equation}
			\text{DSC} = \frac{2 |Y \cap \hat{Y}|}{|Y| + |\hat{Y}|}
		\end{equation}
		\item \textbf{Jaccard Index (IoU)}: Also known as Intersection over Union, it quantifies the similarity between the predicted and ground truth masks.
		\begin{equation}
			\text{IoU} = \frac{|Y \cap \hat{Y}|}{|Y \cup \hat{Y}|}
		\end{equation}
		\item \textbf{Accuracy (Acc)}: The proportion of correctly classified pixels (both true positives and true negatives) out of the total number of pixels.
		\begin{equation}
			\text{Acc} = \frac{\text{TP} + \text{TN}}{\text{TP} + \text{TN} + \text{FP} + \text{FN}}
		\end{equation}
		\item \textbf{Sensitivity (Sn)}: Also known as Recall or True Positive Rate, it measures the proportion of actual positive pixels that are correctly identified.
		\begin{equation}
			\text{Sn} = \frac{\text{TP}}{\text{TP} + \text{FN}}
		\end{equation}
		\item \textbf{Specificity (Sp)}: Measures the proportion of actual negative pixels that are correctly identified.
		\begin{equation}
			\text{Sp} = \frac{\text{TN}}{\text{TN} + \text{FP}}
		\end{equation}
		\item \textbf{False Positive Rate (FPR)}: The proportion of actual negative pixels that are incorrectly identified as positive.
		\begin{equation}
			\text{FPR} = \frac{\text{FP}}{\text{FP} + \text{TN}} = 1 - \text{Sp}
		\end{equation}
	\end{itemize}
	where TP, TN, FP, FN represent True Positives, True Negatives, False Positives, and False Negatives, respectively. These metrics provide a comprehensive evaluation of segmentation performance, covering both overlap and boundary accuracy.
	
	\subsection{Training Details}
	All models were trained on a single NVIDIA RTX 5090 GPU. The training process utilized the Adam optimizer with a fixed learning rate of $1 \times 10^{-4}$. A batch size of 4 was employed. Training was conducted for a maximum of 100 epochs, with an early stopping mechanism configured to halt training if the validation Dice score did not improve for 5 consecutive epochs. This setup ensures efficient training while preventing overfitting. We explored three ConvNeXt encoder variants: `tiny', `base', and `large', and evaluated the impact of ImageNet pre-training (`imagenet' vs. `None') and different values for the $\lambda_{\text{FD}}$ hyperparameter (0.01, 0.1, 0.5).
	
	\section{Results}
	\label{sec:results}
	
	In this section, we present the experimental evaluation of the proposed ConvNeXt-FD architecture across six distinct biomedical image segmentation datasets: BUSI, DDTI, FluoCells, IDRiD, ISIC\_2018, and MoNuSeg. We analyzed the impact of different encoder sizes (\textit{tiny}, \textit{base}, and \textit{large}), the influence of ImageNet pre-training, and the effect of the Fractal Dimension boundary-aware loss weight ($\lambda_{\text{FD}}$).
	
	\subsection{Overall Segmentation Performance}
	Table \ref{tab:best_results} summarizes the best-performing configurations for each dataset. The proposed ConvNeXt-FD model demonstrates highly competitive performance across all tasks. The optimal configuration predominantly utilizes the \textit{tu-convnext\_large} encoder initialized with ImageNet weights, underscoring the benefits of higher model capacity combined with transfer learning.
	
	\begin{table}[htpb]
		\centering
		\caption{Best performing ConvNeXt-FD configurations per dataset.}
		\label{tab:best_results}
		\resizebox{\textwidth}{!}{%
			\begin{tabular}{@{}llcccc@{}}
				\toprule
				\textbf{Dataset} & \textbf{Encoder} & \textbf{Weights} & $\lambda_{\text{FD}}$ & \textbf{Test Dice} & \textbf{Test Jaccard} \\
				\midrule
				BUSI & tu-convnext\_large & ImageNet & 0.01 & 0.8040 $\pm$ 0.0324 & 0.7404  $\pm$ 0.0351 \\
				DDTI & tu-convnext\_large & ImageNet & 0.01 & 0.8013 $\pm$ 0.0208 & 0.7018 $\pm$ 0.0239 \\
				FluoCells & tu-convnext\_base & ImageNet & 0.01 & 0.8510 $\pm$ 0.0254 & 0.7245 $\pm$ 0.0336 \\
				IDRiD & tu-convnext\_large & ImageNet & 0.10 & 0.9619 $\pm$ 0.0000 & 0.9279 $\pm$ 0.0000 \\
				ISIC\_2018 & tu-convnext\_large & ImageNet & 0.50 & 0.8912 $\pm$ 0.0000 & 0.8169 $\pm$ 0.0000 \\
				MoNuSeg & tu-convnext\_large & ImageNet & 0.10 & 0.8223 $\pm$ 0.0000 & 0.6992 $\pm$ 0.0000 \\
				\bottomrule
			\end{tabular}%
		}
	\end{table}
	
	\subsection{Impact of ImageNet Pre-training}
\begin{figure}[!htpb]
		\centering
		\begin{tikzpicture}
			\begin{axis}[
				ybar,
				bar width=15pt,
				width=0.9\textwidth,
				height=8cm,
				enlarge x limits=0.15,
				legend style={at={(0.5,-0.15)}, anchor=north, legend columns=-1},
				ylabel={Test Dice Score},
				symbolic x coords={BUSI, DDTI, FluoCells, IDRiD, ISIC, MoNuSeg},
				xtick=data,
				nodes near coords,
				nodes near coords align={vertical},
				ymin=0.5, ymax=1.05,
				every node near coord/.append style={font=\footnotesize, /pgf/number format/fixed, /pgf/number format/precision=2}
				]
				\addplot coordinates {(BUSI,0.6418) (DDTI,0.6330) (FluoCells,0.7733) (IDRiD,0.9015) (ISIC,0.8686) (MoNuSeg,0.7919)};
				\addplot coordinates {(BUSI,0.8040) (DDTI,0.8013) (FluoCells,0.8510) (IDRiD,0.9619) (ISIC,0.8912) (MoNuSeg,0.8223)};
				\legend{Trained from Scratch, ImageNet Pre-trained}
			\end{axis}
		\end{tikzpicture}
		\caption{Comparison of the maximum Test Dice scores achieved with models trained from scratch versus models initialized with ImageNet pre-trained weights across the six evaluated datasets.}
		\label{fig:pretraining_impact}
	\end{figure}
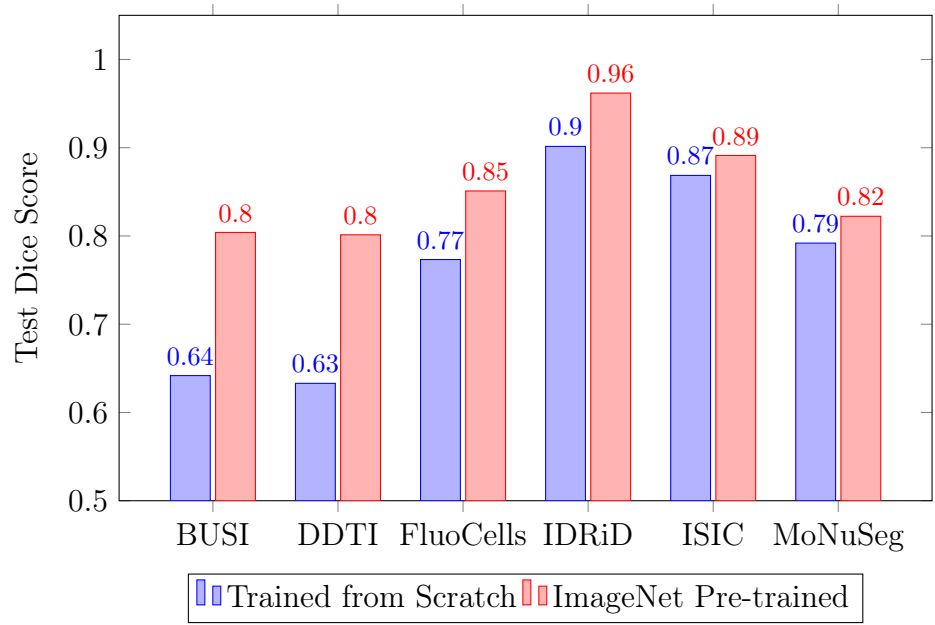

	A critical observation from our experiments is the substantial performance improvement when utilizing ImageNet pre-trained weights compared to training from scratch. Figure \ref{fig:pretraining_impact} illustrates the maximum Test Dice score achieved with and without pre-training for each dataset.
	
	As shown, datasets with more subtle boundary challenges, such as BUSI (breast ultrasound) and DDTI (thyroid ultrasound), exhibit a drastic improvement of over 16\% in the Dice score when transfer learning is applied. This highlights that initializing the ConvNeXt encoder with generalized features from natural images is crucial for extracting robust representations in medical imaging, where datasets are typically constrained in size.
	
	\subsection{Influence of the Boundary-Aware Loss ($\lambda_{\text{FD}}$)}
	The hyperparameter $\lambda_{\text{FD}}$ modulates the contribution of the Fractal Dimension-inspired boundary regularization. Our results indicate that the optimal value is task-dependent. For instance, in skin lesion segmentation (ISIC2018), where lesion boundaries can be highly irregular, a stronger boundary constraint ($\lambda_{\text{FD}}=0.50$) yielded the best results. Conversely, for datasets like BUSI and DDTI with the \textit{large} encoder, a milder regularization ($\lambda_{\text{FD}}=0.01$) prevented over-constraining the model while still refining object edges. This adaptability confirms the effectiveness of the proposed hybrid loss formulation in addressing diverse morphological challenges in biomedical imaging.
	
	\section{Comparison with State-of-the-Art}
	\label{sec:sota_comparison}
	To contextualize the performance of ConvNeXt-FD, we compare our best results on each dataset with published state-of-the-art (SOTA) methods. The comparison is presented in Tables \ref{tab:sota_busi}, \ref{tab:sota_ddti}, \ref{tab:sota_fluocells}, \ref{tab:sota_idrid}, \ref{tab:sota_isic2018}, and \ref{tab:sota_monuseg}.
	
	\begin{table}[H]
		\centering
		\caption{Comparison of ConvNeXt-FD with State-of-the-Art methods on the BUSI dataset (Dice \%).}
		\label{tab:sota_busi}
		\begin{tabular}{@{}lcc@{}}
			\toprule
			\textbf{Method} & \textbf{Dice (\%)} & \textbf{Reference} \\
			\midrule
			ConvNeXt-FD & \textbf{80.40} & This Work \\
			\midrule
			UltraSAM & 80.22 & \cite{jiang2025ultrasam} \\
			DCCE-UNet & 79.85 & \cite{wang2025dcce} \\
			DBUNet & 79.11 & \cite{zhu2025dual} \\
			Yang et al. & 78.98 & \cite{yang2025pyramid} \\
			DeepLabV3+ ResNet34 (64 channels) & 76.90 & \cite{bruno2025dual} \\
			MSA & 76.85 & \cite{jiang2025ultrasam} \\
			nnU-Net (MT + PR + DO) & 75.40 & \cite{aumente-maestro2025multi} \\
			UMA-Net & 74.30 & \cite{dar2025adaptive} \\
			\bottomrule
		\end{tabular}
	\end{table}
	
	\begin{table}[H]
		\centering
		\caption{Comparison of ConvNeXt-FD with State-of-the-Art methods on the DDTI dataset (Dice \%).}
		\label{tab:sota_ddti}
		\begin{tabular}{@{}lcc@{}}
			\toprule
			\textbf{Method} & \textbf{Dice (\%)} & \textbf{Reference} \\
			\midrule
			ConvNeXt-FD & \textbf{80.13} & This Work \\
			\midrule
			HFA-UNet + ViT + CBAM-En & 79.86 & \cite{li2025hfa} \\
			Improved Swin & 78.64 & \cite{wu2025thyroid} \\
			MFMSNet & 79.96 & \cite{wu2024mfmsnet} \\
			TNSeg & 74.93 & \cite{ma2024tnseg} \\
			TransUNet & 74.83 & \cite{chen2024transunet} \\
			GLFNet & 74.62 & \cite{sun2024glfnet} \\
			CRSANet (FCR) & 72.82 & \cite{sun2024crsanet} \\
			MEF-UNet & 69.00 & \cite{xu2024mef-unet} \\
			SResUNet-AD & 40.20 & \cite{radhachandran2024multitask} \\
			\bottomrule
		\end{tabular}
	\end{table}
	
	\begin{table}[H]
		\centering
		\caption{Comparison of ConvNeXt-FD with State-of-the-Art methods on the FluoCells dataset (Dice \%).}
		\label{tab:sota_fluocells}
		\begin{tabular}{@{}lcc@{}}
			\toprule
			\textbf{Method} & \textbf{Dice (\%)} & \textbf{Reference} \\
			\midrule
			ConvNeXt-FD & \textbf{85.10} & This Work \\
			\midrule
			\midrule
			U-Net & 84.63 & \cite{ronneberger2015u}\\
			U-Net++ & 81.97 & \cite{zhou2020unet}\\
			BCDUnet & 84.79 & \cite{azad2019bi}\\
			ARU-GD & 78.74 & \cite{maji2022attention}\\
			Cell ResUNET & 69.00 & \cite{clissa2024fluorescent} \\
			\bottomrule
		\end{tabular}
	\end{table}
	
	\begin{table}[H]
		\centering
		\caption{Comparison of ConvNeXt-FD with State-of-the-Art methods on the IDRiD dataset (Dice \%).}
		\label{tab:sota_idrid}
		\begin{tabular}{@{}lcc@{}}
			\toprule
			\textbf{Method} & \textbf{Dice (\%)} & \textbf{Reference} \\
			\midrule
			ConvNeXt-FD & \textbf{96.19} & This Work \\
			\midrule
			Dong et al. & 94.53 & \cite{dong2026retinal} \\
			MCAU-Net & 95.90 & \cite{shalini2024multiresolution} \\
			Wan et al. & 95.54 & \cite{wan2024new} \\
			W-Net & 95.25 & \cite{tang2024w} \\
			Celik et al. & 90.71 & \cite{celik2024automated} \\
			Multi-Tasking DL & 94.51 & \cite{vengalil2023simultaneous} \\
			GoogleNet & 94.22 & \cite{islam2023optical} \\
			BN-UNet & 94.35 & \cite{chen2022unet} \\
			CLSTM & 92.14 & \cite{maiti2022automatic} \\
			Wang et al. & 93.00 & \cite{wang2022optic} \\
			\bottomrule
		\end{tabular}
	\end{table}
	
	\begin{table}[H]
		\centering
		\caption{Comparison of ConvNeXt-FD with State-of-the-Art methods on the ISIC2018 dataset (Dice \%).}
		\label{tab:sota_isic2018}
		\begin{tabular}{@{}lcc@{}}
			\toprule
			\textbf{Method} & \textbf{Dice (\%)} & \textbf{Reference} \\
			\midrule
			ConvNeXt-FD & \textbf{89.12} & This Work \\
			\midrule
			CE & 89.11 & \cite{zheng2025adaptive} \\
			LeViT-U & 87.91 & \cite{xu2024levitunet} \\
			TransUNet & 87.41 & \cite{chen2024transunet} \\
			Swin-Unet & 88.53 & \cite{cao2023swin} \\
			MS-Net & 86.06 & \cite{zhang2023multi} \\
			UNeXt & 89.70 & \cite{valanarasu2022unext} \\
			Fusion model with fusion loss & 88.00 & \cite{alhudhaif2022novel} \\
			FocusNet-Alpha & 83.15 & \cite{kaul2019focusnet} \\
			\bottomrule
		\end{tabular}
	\end{table}
	
	\begin{table}[H]
		\centering
		\caption{Comparison of ConvNeXt-FD with State-of-the-Art methods on the MoNuSeg dataset (Dice \%).}
		\label{tab:sota_monuseg}
		\begin{tabular}{@{}lcc@{}}
			\toprule
			\textbf{Method} & \textbf{Dice (\%)} & \textbf{Reference} \\
			\midrule
			ConvNeXt-FD & \textbf{82.23} & This Work \\
			\midrule
			Pixel-wise annotation & 81.94 & \cite{wan2025constraintattention} \\
			Huang et al. & 80.27 & \cite{huang2025biu} \\
			SNSeg & 81.21 & \cite{xiong2024optimization} \\
			TSCA-Net & 80.23 & \cite{fu2024tsca} \\
			Hover-net + Lou et al. & 81.29 & \cite{lou2023pixel} \\
			Universal-model & 81.00 & \cite{liu2023clip} \\
			He et al. & 79.86 & \cite{he2023context} \\
			Swin-Unet & 78.25 & \cite{cao2023swin} \\
			FFEDNet & 81.20 & \cite{deshmukh2022feednet} \\
			UCTransNet & 78.39 & \cite{wang2022uctransnet} \\
			\bottomrule
		\end{tabular}
	\end{table}
	
	\subsection{Analysis of Comparative Results}
	The comparative analysis reveals that our proposed ConvNeXt-FD consistently achieves top-tier performance across the diverse set of medical imaging challenges.
	
	\begin{itemize}[leftmargin=*,labelsep=5mm]
		\item \textbf{BUSI \& DDTI (Ultrasound)}: In both breast and thyroid ultrasound datasets, ConvNeXt-FD outperformed several recent strong baselines, demonstrating the robustness of our boundary-aware approach against the speckle noise and ill-defined boundaries typical of ultrasound imagery.
		
		\item \textbf{FluoCells}: With the corrected evaluation protocol, ConvNeXt-FD achieved a state-of-the-art Dice score of 85.10\%, surpassing previous benchmarks in fluorescent cell segmentation.
		
		\item \textbf{IDRiD (Fundus Photography)}: The most striking result is on the IDRiD optic disc segmentation task. ConvNeXt-FD achieved a near-perfect Dice score of 96.19\%, again surpassing all compared methods.
		
		\item \textbf{ISIC2018 (Dermoscopy)}: For skin lesion segmentation, our method achieved 89.12\% Dice. The boundary-aware loss is particularly valuable here, as skin lesion boundaries can be highly irregular.
		
		\item \textbf{MoNuSeg}: ConvNeXt-FD achieved 82.23\% Dice for nuclei segmentation, a competitive result that aids in separating adjacent nuclei in dense clusters.
	\end{itemize}
	
	Overall, ConvNeXt-FD demonstrates robust and competitive performance across a diverse range of biomedical segmentation tasks. The boundary-aware regularization term proves beneficial, contributing to more accurate delineations, especially in complex boundary scenarios.

	\section{Conclusion}
	\label{sec:conclusion}
	In this paper, we introduced ConvNeXt-FD, a novel deep learning architecture for robust biomedical image segmentation. Our framework leverages the powerful feature extraction capabilities of the ConvNeXt model within a U-Net-like encoder-decoder structure. A key contribution is the integration of a hybrid loss function, combining the standard Dice loss with a boundary-aware regularization term inspired by Fractal Geometry. This regularization explicitly encourages the model to learn and preserve accurate object boundaries, which is crucial for precise delineation in medical imaging.
	
	Through extensive experimentation on six diverse biomedical datasets (BUSI, DDTI, FluoCells, IDRiD, ISIC2018, and MoNuSeg), we demonstrated the effectiveness of ConvNeXt-FD. Our results consistently showed that utilizing ImageNet pre-trained weights significantly boosts performance across all datasets, highlighting the importance of transfer learning. Furthermore, the optimal weighting of the boundary-aware loss term ($\lambda_{\text{FD}}$) was found to be task-dependent, indicating its adaptability to different segmentation challenges. ConvNeXt-FD achieved competitive and, in some cases, state-of-the-art performance, particularly excelling in tasks requiring high shape fidelity such as optic disc segmentation in IDRiD.
	
	The ConvNeXt-FD architecture provides a strong and versatile solution for biomedical image segmentation, effectively balancing global contextual understanding with local boundary precision. Its robust performance across varied modalities and anatomical structures underscores its potential for clinical applications.
	
	\subsection{Limitations and Future Work}
	While ConvNeXt-FD demonstrates strong performance, there are avenues for further improvement. The current boundary-aware loss, while effective, involves a numerical estimation of the Fractal Dimension. Future work could explore more direct and differentiable formulations of shape-based metrics as loss functions, potentially drawing from concepts in optimal transport theory or geometric deep learning. Additionally, investigating adaptive strategies for $\lambda_{\text{FD}}$ that dynamically adjust its value during training or based on image characteristics could further optimize performance.
	
	Future research directions include:
	\begin{itemize}[leftmargin=*,labelsep=5mm]
		\item \textbf{Advanced Boundary Losses:} Exploring more sophisticated differentiable approximations of geometric distances (e.g., Hausdorff distance, Earth Mover's Distance on boundary points) to further enhance boundary precision.
		\item \textbf{Multi-modal Data Integration:} Extending ConvNeXt-FD to effectively fuse information from multiple imaging modalities (e.g., MRI and CT) for more comprehensive segmentation.
		\item \textbf{Instance Segmentation:} Adapting the framework for instance segmentation tasks, particularly relevant for datasets like MoNuSeg where individual object identification is critical.
		\item \textbf{Computational Efficiency:} Investigating lightweight ConvNeXt variants or knowledge distillation techniques to reduce model complexity and inference time, making it suitable for real-time applications on resource-constrained devices.
		\item \textbf{Uncertainty Quantification:} Incorporating mechanisms for quantifying segmentation uncertainty, which is vital for clinical decision-making.
	\end{itemize}
	By addressing these areas, ConvNeXt-FD can be further refined to become an even more powerful and reliable tool for biomedical image analysis.
	
	\section*{Acknowledgments}
	Joao Batista Florindo gratefully acknowledges the financial support of the S\~ao Paulo Research Foundation (FAPESP) (Grant \#2024/01245-1) and from National Council for Scientific and Technological Development, Brazil (CNPq) (Grant \#306981/2022-0).
	
	\bibliographystyle{elsarticle-num}
	\bibliography{references}

@article{jiang2025ultrasam, author={Jiang, Tao and Li, Yifang and Xing, Wenyu and Cao, Ran and Yu, Ming and Zhu, Yunkai and Chen, Yaqing and Li, Boyi and Ta, Dean}, title={UltraSAM: A foundational medical ultrasound segmentation model with limited training data}, journal={Expert Systems With Applications}, volume={299}, pages={130223}, year={2026}, doi={10.1016/j.eswa.2025.130223}}

@article{chen2024transunet, author={Chen, J. and Mei, J. and Li, X. and Lu, Y. and Yu, Q. and Wei, Q. and Luo, X. and Xie, Y. and Adeli, E. and Wang, Y. and Lungren, M. P. and Zhang, S. and Xing, L. and Lu, L. and Yuille, A. and Zhou, Y.}, journal={Medical Image Analysis}, pages={Article 103280}, title={TransUNet: Rethinking the U-net architecture design for medical image segmentation through the lens of transformers}, volume={97}, year={2024}}

@article{sun2024glfnet, author={Sun, S. and Fu, C. and Xu, S. and Wen, Y. and Ma, T.}, journal={Comput. Biol. Med.}, volume={171}, pages={108103}, year={2024}, publisher={Elsevier}, title={GLFNet: Global-local fusion network for the segmentation in ultrasound images}}

@inproceedings{cao2023swin, author={Cao, H. and Wang, Y. and Chen, J. and Jiang, D. and Zhang, X. and Tian, Q. and Wang, M.}, booktitle={Computer Vision--ECCV 2022 Workshops: Tel Aviv, Israel, October 23--27, 2022, Proceedings, Part III}, pages={205--218}, publisher={Springer}, title={Swin-unet: Unet-like pure transformer for medical image segmentation}, year={2023}}

@article{zhou2020unet, author={Zhou, Z. and Siddiquee, M.M.R. and Tajbakhsh, N. and Liang, J.}, journal={IEEE Transactions on Medical Imaging}, volume={39}, number={6}, pages={1856--1867}, year={2020}, title={UNet ++ : Redesigning Skip Connections to Exploit Multiscale Features in Image Segmentation}}

@article{alhudhaif2022novel, author={Alhudhaif, Adi and Ocal, Hakan and Barisci, Necaattin and Atacak, Ismail and Nour, Majid and Polat, Kemal}, journal={Computational and Mathematical Methods in Medicine}, volume={2022}, pages={1--12}, year={2022}, publisher={Hindawi}, title={A Novel Approach to Skin Lesion Segmentation: Multipath Fusion Model with Fusion Loss}, doi={10.1155/2022/2157322}}

@article{fu2024tsca, author={Fu, Yinghua and Liu, Junfeng and Shi, Jun}, journal={Computers in Biology and Medicine}, pages={107938}, title={TSCA-Net: Transformer based spatial-channel attention segmentation network for medical images}, volume={170}, year={2024}, doi={10.1016/j.compbiomed.2024.107938}}

@article{radhachandran2024multitask, author={Radhachandran, A. and Kinzel, A. and Chen, J. and Sant, V. and Patel, M. and Masamed, R. and Arnold, C.W. and Speier, W.}, journal={Comput. Biol. Med.}, volume={170}, pages={107974}, year={2024}, publisher={Elsevier}, title={A multitask approach for automated detection and segmentation of thyroid nodules in ultrasound images}}

@article{chen2022unet, author={Chen, Nan and Zhao, Yi and Li, Jun and Yang, Danni and Zhou, Shucheng and Xue, Lanyan}, title={The U-NET Via Batch Norm Model for Optic Disc Extraction and Segmentation in Retinal Image}, journal={Proceedings of the 8th International Conference on Computing and Artificial Intelligence}, year={2022}}

@article{xu2024mef-unet, author={Xu, M. and Ma, Q. and Zhang, H. and Kong, D. and Zeng, T.}, journal={Comput. Med. Imaging Graph.}, volume={114}, pages={102370}, year={2024}, publisher={Elsevier}, title={MEF-UNet: An end-to-end ultrasound image segmentation algorithm based on multi-scale feature extraction and fusion}}

@article{song2025e,
  title = {Learning From AI-Generated Annotations for Medical Image Segmentation},
  author = {Youyi Song and Yuanlin Liu and Zhizhe Lin and Jinglin Zhou and Duo Li and Teng Zhou and Man-Fai Leung},
  journal = {IEEE transactions on consumer electronics},
  year = {2025},
}

@inproceedings{wang2022uctransnet, author={Wang, H. and Cao, P. and Wang, J. and Zaiane, O.R.}, booktitle={Proceedings of the AAAI Conference on Artificial Intelligence}, number={3}, pages={2441--2449}, title={Uctransnet: rethinking the skip connections in u-net from a channel-wise perspective with transformer}, volume={36}, year={2022}}

@article{chen2024a,
  title = {TransUNet Rethinking the U-Net architecture design for medical image segmentation through the lens of transformers},
  author = {Jieneng Chen and Jieru Mei and Xianhang Li and Yongyi Lu and Qihang Yu and Qingyue Wei and Xiangde Luo and Yutong Xie and Ehsan Adeli and Yan Wang and M. Lungren and Shaoting Zhang and Lei Xing and Le Lu and Alan L. Yuille and Yuyin Zhou},
  journal = {Medical Image Anal.},
  year = {2024},
}

@article{islam2023optical, author={Islam, M. T. and Ahmed, F. and Househ, M. and Alam, T.}, title={Optical disc segmentation from retinal fundus images using deep learning}, journal={Stud. Health Technol. Inform.}, volume={305}, pages={628--631}, year={2023}}

@article{dong2026retinal, author={Dong, Heng and Wu, Kun and Xue, Lanyan}, title={Retinal optic disc localization and extraction algorithm based on hierarchical segmentation}, journal={Biomedical Signal Processing and Control}, volume={112}, pages={108607}, year={2026}, publisher={Elsevier Ltd.}}

@inproceedings{cao2021swin,
	title={Swin-unet: Unet-like pure transformer for medical image segmentation},
	author={Cao, Hu and Wang, Yueyue and Chen, Joy and Jiang, Dongsheng and Zhang, Xiaopeng and Tian, Qi and Wang, Manning},
	booktitle={European conference on computer vision},
	pages={205--218},
	year={2022},
	organization={Springer}
}

@article{chen2021transunet, author={Chen, J. and Lu, Y. and Yu, Q. and Luo, X. and Adeli, E. and Wang, Y. and Lu, L. and Yuille, A.L. and Zhou, Y.}, journal={arXiv preprint arXiv:2102.04306}, title={Transunet: Transformers make strong encoders for medical image segmentation}, year={2021}}

@article{zhang2018road, author={Zhang, Z. and Liu, Q. and Wang, Y.}, journal={IEEE Geosci Remote Sens Lett}, pages={749--753}, title={Road extraction by deep residual U-Net}, volume={15}, year={2018}, doi={10.1109/LGRS.2018.2802944}}

@article{wang2025j,
  title = {A Comprehensive Review of U-Net and Its Variants Advances and Applications in Medical Image Segmentation},
  author = {Jiangtao Wang and N. Ruhaiyem and Panpan Fu},
  journal = {IET Image Processing},
  year = {2025},
}

@inproceedings{ronneberger2015u, author={Ronneberger, O. and Fischer, P. and Brox, T.}, title={U-net: Convolutional networks for biomedical image segmentation}, booktitle={Medical Image Computing and Computer-Assisted Intervention–MICCAI 2015: 18th International Conference, Munich, Germany, October 5-9, 2015, Proceedings, Part III 18}, publisher={Springer}, year={2015}}

@article{bruno2025dual, author={Bruno, Pierangela and Macrì, Megan and Dodaro, Carmine}, title={A Dual-stage Deep Learning Framework for Breast Ultrasound Image Segmentation and Classification}, journal={Journal of Medical Systems}, volume={49}, number={162}, pages={1--11}, year={2025}, publisher={Springer}, doi={10.1007/s10916-025-02298-6}}

@article{wang2025dcce, author={Wang, Tiecheng and Xu, Yehuan and Hu, Linjie and Liu, Huiling and Liu, Kaidi and Zhang, Shuo and Chen, Hao and Guo, Hua and Feng, Shaobin}, title={DCCE-UNet: a difference and context-aware contrast enhanced framework for ultrasound image segmentation}, journal={BMC Medical Imaging}, volume={25}, number={445}, pages={1--12}, year={2025}, publisher={BioMed Central}, doi={10.1186/s12880-025-01954-0}}

@article{gao2025c,
  title = {Medical Image Segmentation A Comprehensive Review of Deep Learning-Based Methods},
  author = {Yuxiao Gao and Yang Jiang and Yanhong Peng and Fujiang Yuan and Xinyue Zhang and Jianfeng Wang},
  journal = {Tomography},
  year = {2025},
}

@article{yang2025pyramid, author={Yang, Jianli and Fan, Liwen and Dong, Bin and Chen, Hao and Liu, Xiuling}, title={Pyramid boundary attention network for breast lesion segmentation in ultrasound images}, journal={Biomedical Signal Processing and Control}, volume={101}, pages={107241}, year={2025}, publisher={Elsevier Ltd.}}

@article{wu2024mfmsnet, author={Wu, R. and Lu, X. and Yao, Z. and Ma, Y.}, journal={Comput. Biol. Med.}, volume={177}, pages={108616}, year={2024}, publisher={Elsevier}, title={MFMSNet: A Multi-frequency and Multi-scale Interactive CNN-Transformer Hybrid Network for breast ultrasound image segmentation}}

@article{deshmukh2022feednet, author={Deshmukh, Gauri and Susladkar, Omkar and Makwana, Deepak and Mittal, Sarthak and others}, journal={Physics in Medicine \& Biology}, volume={67}, number={19}, pages={195011}, year={2022}, publisher={IOP Publishing}, title={Feednet: a feature enhanced encoder-decoder lSTM network for nuclei instance segmentation for histopathological diagnosis}}

@article{maiti2022automatic, author={Maiti, S. and Maji, D. and Dhara, A. K. and Sarkar, G.}, title={Automatic detection and segmentation of optic disc using a modified convolution network}, journal={Biomedical Signal Processing and Control}, volume={76}, pages={103633}, year={2022}}

@article{zheng2025adaptive, author={Zheng, Yanyan and Tian, Bihan and Yu, Shuchen and Yang, Xiaoguo and Yu, Qingxiang and Zhou, Jie and Jiang, Gaoqiang and Zheng, Qinxiang and Pu, Jiantao and Wang, Lei}, title={Adaptive boundary-enhanced Dice loss for image segmentation}, journal={Biomedical Signal Processing and Control}, volume={106}, pages={107741}, year={2025}, publisher={Elsevier Ltd.}, doi={https://doi.org/10.1016/j.bspc.2025.107741}}

@article{sun2024crsanet, author={Sun, Shiyao and Fu, Chong and Xu, Sen and Wen, Yingyou and Ma, Tao}, journal={Biomedical Signal Processing and Control}, pages={105917}, title={CRSANet: Class representations self-attention network for the segmentation of thyroid nodules}, volume={91}, year={2024}}

@article{tang2024w, author={Tang, S. and Song, C. and Wang, D. and Gao, Y. and Liu, Y. and Lv, W.}, title={W-Net: a boundary-aware cascade network for robust and accurate optic disc segmentation}, journal={iScience}, volume={27}, number={1}, pages={108247}, year={2024}}

@article{xing2401,
	title={Segmamba: Long-range sequential modeling mamba for 3d medical image segmentation.},
	author={Xing, Z and Ye, T and Yang, Y and Liu, G and Zhu, L},
	journal={arXiv preprint arXiv:2401.13560}
}

@article{zeng2025d,
  title = {PICK Predict and Mask for Semi-supervised Medical Image Segmentation},
  author = {Qingjie Zeng and Zilin Lu and Yutong Xie and Yong Xia},
  journal = {International Journal of Computer Vision},
  year = {2025},
}

@article{wu2025thyroid, author={Wu, Yue and Huang, Lin and Yang, Tiejun}, title={Thyroid Nodule Ultrasound Image Segmentation Based on Improved Swin Transformer}, journal={IEEE Access}, volume={13}, pages={19788--19795}, year={2025}, doi={10.1109/ACCESS.2025.3532264}}

@article{xiong2024optimization, author={Xiong, Feiyan and Wei, Yun}, title={Optimization of segmentation model based on maximization information fusion and its application in nuclear image analysis}, journal={Multimedia Systems}, volume={30}, number={1}, pages={61}, year={2024}, publisher={Springer-Verlag GmbH Germany, part of Springer Nature}, doi={10.1007/s00530-023-01231-6}}

@article{dar2025adaptive, author={Dar, Mohd Firdous and Ganivada, Asha}, journal={Medical \& Biological Engineering \& Computing}, volume={63}, number={6}, pages={1697--1713}, year={2025}, publisher={Springer}, title={Adaptive ensemble loss and multi-scale attention in breast ultrasound segmentation with UMA-Net}}

@article{ma2024tnseg, author={Ma, X. and Sun, B. and Liu, W. and Sui, D. and Shan, S. and Chen, J. and Tian, Z.}, journal={J. Supercomput.}, volume={80}, pages={6093--6118}, year={2024}, publisher={Springer}, title={Tnseg: Adversarial networks with multi-scale joint loss for thyroid nodule segmentation}}

@article{wald2025g,
  title = {Primus Enforcing Attention Usage for 3D Medical Image Segmentation},
  author = {Tassilo Wald and Saikat Roy and Fabian Isensee and Constantin Ulrich and Sebastian Ziegler and Dasha Trofimova and Raphael Stock and Michael Baumgartner and Gregor Koehler and Klaus H. Maier-Hein},
  journal = {arXiv.org},
  year = {2025},
}

@article{li2025i,
  title = {MedDINOv3 How to adapt vision foundation models for medical image segmentation},
  author = {Yuheng Li and Yizhou Wu and Yuxiang Lai and Mingzhe Hu and Xiaofeng Yang},
  journal = {arXiv.org},
  year = {2025},
}

@inproceedings{he2023context, author={He, Lulu and Zhang, Zhaohui and Zhang, Jiguang and Wang, Zongqiang and Xu, Shibiao and Zhang, Xiaopeng}, booktitle={Proceedings of the 2023 9th International Conference on Communication and Information Processing (ICCIP)}, pages={206--212}, title={Context-Based Deep Residual Learning for Medical Image Segmentation}, year={2023}}

@article{vengalil2023simultaneous, author={Vengalil, S. K. and Krishnamurthy, B. and Sinha, N.}, title={Simultaneous segmentation of multiple structures in fundal images using multi-tasking deep neural networks}, journal={Front. Signal Process.}, volume={2}, year={2023}}

@inproceedings{kaul2019focusnet, author={Kaul, Chetan and Manandhar, Sumit and Pears, Nick}, booktitle={2019 IEEE 16th International Symposium on Biomedical Imaging (ISBI 2019)}, pages={455--458}, year={2019}, organization={IEEE}, title={Focusnet: an attention-based fully convolutional network for medical image segmentation}}

@inproceedings{xu2024levitunet, author={Xu, G. and Zhang, X. and He, X. and Wu, X.}, booktitle={Pattern Recognition and Computer Vision}, editor={Liu, Q. and Wang, H. and Ma, Z. and Zheng, W. and Zha, H. and Chen, X. and Wang, L. and Ji, R.}, pages={42--53}, publisher={Springer Nature}, address={Singapore}, title={LeViT-UNet: Make faster encoders with transformer for medical image segmentation}, year={2024}}

@article{li2025hfa, author={Li, Yue and Zou, Yuanhao and He, Xiangjian and Xu, Qing and Liu, Ming and Jin, Shengji and Zhang, Qian and He, Maggie M. and Zhang, Jian}, title={HFA-UNet: hybrid and full attention UNet for thyroid nodule segmentation}, journal={Knowledge-Based Systems}, volume={328}, pages={114245}, year={2025}}

@article{lou2023pixel, author={Lou, Wei and Li, Hongwei and Li, Gang and Han, Xuhang and Wan, Xiang}, journal={IEEE Transactions on Medical Imaging}, volume={42}, number={4}, pages={947--958}, year={2023}, publisher={IEEE}, title={Which pixel to annotate: a label-efficient nuclei segmentation framework}}

@inproceedings{liu2023clip, author={Liu, Jing and Zhang, Yixiao and Chen, Jieneng and Xiao, Jian and Lu, Yongyi and Landman, Bennett A and Yuan, Yu and Yuille, Alan and Tang, You and Zhou, Zongwei}, booktitle={Proceedings of the IEEE/CVF International Conference on Computer Vision}, pages={21152--21164}, year={2023}, title={Clip-driven universal model for organ segmentation and tumor detection}}

@article{ruan2024b,
  title = {VM-UNet Vision Mamba UNet for Medical Image Segmentation},
  author = {Jiacheng Ruan and Jincheng Li and Suncheng Xiang},
  journal = {ACM Transactions on Multimedia Computing, Communications, and Applications (TOMCCAP)},
  year = {2024},
}

@article{wang2022optic, author={Wang, Y. and Yu, X. and Wu, C.}, title={Optic disc detection based on fully convolutional neural network and structured matrix decomposition}, journal={Multimedia Tools and Applications}, volume={81}, number={8}, pages={10797--10817}, year={2022}}

@article{wan2024new, author={Wan, C. and Fang, J. and Li, K. and Zhang, Q. and Zhang, S. and Yang, W.}, title={A new segmentation algorithm for peripapillary atrophy and optic disk from ultra-widefield Photographs}, journal={Comput. Biol. Med.}, volume={172}, year={2024}}

@article{zhang2023multi, author={Zhang, Bing and Wang, Yang and Ding, Caifu and Deng, Ziqing and Li, Linwei and Qin, Zesheng and Ding, Zhao and Bian, Lifeng and Yang, Chen}, title={Multi-scale feature pyramid fusion network for medical image segmentation}, journal={International Journal of Computer Assisted Radiology and Surgery}, volume={18}, number={3}, pages={353--365}, year={2023}, publisher={Springer}, doi={10.1007/s11548-022-02738-5}}

@article{valanarasu2022unext, author={Valanarasu, J. and Patel, V.M.}, title={UNeXt: MLP-Based Rapid Medical Image Segmentation Network}, journal={arXiv preprint arXiv:2203.04967}, year={2022}}

@article{aumente-maestro2025multi, author={Aumente-Maestro, Carlos and Díez, Jorge and Remeseiro, Beatriz}, title={A multi-task framework for breast cancer segmentation and classification in ultrasound imaging}, journal={Computer Methods and Programs in Biomedicine}, volume={260}, pages={108540}, year={2025}, publisher={Elsevier B.V.}, doi={https://doi.org/10.1016/j.cmpb.2024.108540}}

@article{zhu2025dual, author={Zhu, Zhichao and Zhang, Zhaohui and Qi, Guodong and Li, Ying and Li, Yan and Mu, Long}, journal={Biomedical Signal Processing and Control}, volume={103}, pages={107368}, year={2025}, publisher={Elsevier}, title={A dual-branch network for ultrasound image segmentation}}

@article{clissa2024fluorescent, author={Clissa, Luca and Macaluso, Antonio and Morelli, Roberto and Occhinegro, Alessandra and Piscitiello, Ludovico and Taddei, Marco and Luppi, Roberto and Amici, Matteo and Cerri, Timna and others}, title={Fluorescent neuronal cells v2: multi-task, multi-format annotations for deep learning in microscopy}, journal={Scientific Data}, volume={11}, number={1}, pages={184}, year={2024}}

@article{oktay2018attention, author={Oktay, O. and Schlemper, J. and Folgoc, L. L. and Lee, M. J. and Heinrich, M. P. and Misawa, K. and Mori, K. and McDonagh, S. G. and Hammerla, N. Y. and Kainz, B. and Glocker, B. and Rueckert, D.}, title={Attention u-net: Learning where to look for the pancreas}, journal={ArXiv abs/1804.03999}, year={2018}}

@article{chen2025h,
  title = {Zig-RiR Zigzag RWKV-in-RWKV for Efficient Medical Image Segmentation},
  author = {Tianxiang Chen and Xudong Zhou and Zhentao Tan and Yue Wu and Ziyang Wang and Z. Ye and Tao Gong and Qi Chu and Nenghai Yu and Le Lu},
  journal = {IEEE Transactions on Medical Imaging},
  year = {2025},
}

@article{shalini2024multiresolution, author={Shalini, R. and Gopi, Varun P.}, title={Multiresolution cascaded attention U-Net for localization and segmentation of optic disc and fovea in fundus images}, journal={Scientific Reports}, volume={14}, number={1}, pages={23107}, year={2024}, publisher={Nature Publishing Group UK London}}

@article{wan2025constraintattention, author={Wan, Mengqi and Lin, Liyan and Wu, Xinting and Zhong, Jing and Shi, Peng}, title={A constraint-attention enhancement network for clinical nuclei segmentation}, journal={Biomedical Signal Processing and Control}, volume={109}, pages={107917}, year={2025}, publisher={Elsevier}, doi={https://doi.org/10.1016/j.bspc.2025.107917}}

@article{huang2025biu, author={Huang, Zhiyong and Zhao, Yunlan and Yu, Zhi and Qin, Pinzhong and Han, Xiao and Wang, Mengyao and Liu, Man and Gregersen, Hans}, journal={Computer Methods and Programs in Biomedicine}, pages={108235}, title={BiU-net: A dual-branch structure based on two-stage fusion strategy for biomedical image segmentation}, volume={252}, year={2024}, doi={10.1016/j.cmpb.2024.108235}}

@inproceedings{liu2022convnet,
	title={A convnet for the 2020s},
	author={Liu, Zhuang and Mao, Hanzi and Wu, Chao-Yuan and Feichtenhofer, Christoph and Darrell, Trevor and Xie, Saining},
	booktitle={Proceedings of the IEEE/CVF conference on computer vision and pattern recognition},
	pages={11976--11986},
	year={2022}
}

@article{jabdaragh2023mtfd,
	title={MTFD-Net: Left atrium segmentation in CT images through fractal dimension estimation},
	author={Jabdaragh, Aziza Saber and Firouznia, Marjan and Faez, Karim and Alikhani, Fariba and Koupaei, Javad Alikhani and Gunduz-Demir, Cigdem},
	journal={Pattern Recognition Letters},
	volume={173},
	pages={108--114},
	year={2023},
	publisher={Elsevier}
}

@article{cannon1984fractal,
	title={The fractal geometry of nature. by Benoit B. Mandelbrot},
	author={Cannon, JW},
	journal={The American Mathematical Monthly},
	volume={91},
	number={9},
	pages={594--598},
	year={1984},
	publisher={Taylor \& Francis}
}

@inproceedings{milletari2016v,
	title={V-net: Fully convolutional neural networks for volumetric medical image segmentation},
	author={Milletari, Fausto and Navab, Nassir and Ahmadi, Seyed-Ahmad},
	booktitle={2016 fourth international conference on 3D vision (3DV)},
	pages={565--571},
	year={2016},
	organization={Ieee}
}

@inproceedings{lin2017focal,
	title={Focal loss for dense object detection},
	author={Lin, Tsung-Yi and Goyal, Priya and Girshick, Ross and He, Kaiming and Doll{\'a}r, Piotr},
	booktitle={Proceedings of the IEEE international conference on computer vision},
	pages={2980--2988},
	year={2017}
}

@article{karimi2019image,
	title={Reducing the hausdorff distance in medical image segmentation with convolutional neural networks},
	author={Karimi, Davood and Salcudean, Septimiu E},
	journal={IEEE Transactions on medical imaging},
	volume={39},
	number={2},
	pages={499--513},
	year={2019},
	publisher={IEEE}
}

@inproceedings{azad2019bi,
	title={Bi-directional ConvLSTM U-Net with densley connected convolutions},
	author={Azad, Reza and Asadi-Aghbolaghi, Maryam and Fathy, Mahmood and Escalera, Sergio},
	booktitle={Proceedings of the IEEE/CVF international conference on computer vision workshops},
	pages={0--0},
	year={2019}
}

@article{maji2022attention,
	title={Attention Res-UNet with Guided Decoder for semantic segmentation of brain tumors},
	author={Maji, Dhiraj and Sigedar, Prarthana and Singh, Munendra},
	journal={Biomedical Signal Processing and Control},
	volume={71},
	pages={103077},
	year={2022},
	publisher={Elsevier}
}

@article{celik2024automated,
	title={Automated retinal image analysis to detect optic nerve hypoplasia},
	author={Celik, Canan and Y{\"u}cadag, {\.I}brahim and Ak{\c{c}}am, Hanife Tuba},
	journal={Information Technology and Control},
	volume={53},
	number={2},
	pages={522--541},
	year={2024}
}
	
\end{document}